\documentclass[11pt]{article}

\usepackage[final]{acl}
\usepackage{times}
\usepackage{latexsym}
\usepackage[T1]{fontenc}
\usepackage[utf8]{inputenc}
\usepackage{microtype}
\usepackage{inconsolata}
\usepackage{graphicx}
\usepackage{booktabs}
\usepackage{multirow}
\usepackage{amsmath}
\usepackage{amssymb}
\usepackage{enumitem}
\usepackage{array}
\usepackage{float}
\usepackage{placeins}

\title{ContractEval: Query-Conditioned Execution Matching for Procedural Instruction Conformance}

\author{
\textbf{Praphul Singh}\textsuperscript{\(\dagger\)},
\textbf{Shanu Kumar}\textsuperscript{\(\ddagger\)},
\textbf{Akshat Agarwal}\textsuperscript{\(\dagger\)},
\textbf{Ganesh Kumar}\textsuperscript{\(\dagger\)}
\\
{\normalfont
\textsuperscript{\(\dagger\)}Oracle Health AI
\quad
\textsuperscript{\(\ddagger\)}MBZUAI}
}

\begin{document}
\maketitle
\begin{abstract}
As LLM agents move from answering questions to carrying out procedures,
failures can be unwarranted rather than visibly wrong: the final response looks
acceptable even though the system skipped the check, branch, dependency, or
invariant that made the answer justified. Output-only evaluation sees the
answer, and trace-aware judging sees activity, but neither identifies which
obligations were active for the query. We introduce \textsc{ContractEval}, a
diagnostic framework for making those active obligations explicit. It represents
procedural instructions as query-active obligations and matches them against
response or trace evidence, turning omissions, wrong branches, ordering errors,
extra actions, invariant breaches, and output-contract violations into distinct
conformance failures. On a controlled suite of audited procedural contracts,
output-only and trace-aware LLM judges miss many injected structural failures;
under gold expected and observed graphs, ContractEval detects and localizes all
of them. LLM-backed extraction preserves much of this signal but remains
calibration-sensitive. ContractEval is therefore not a compliance guarantee; it
makes procedural conformance auditable rather than implicit in final-answer
quality.
\end{abstract}

\begin{figure*}[t]
  \centering
  \includegraphics[width=0.90\linewidth]{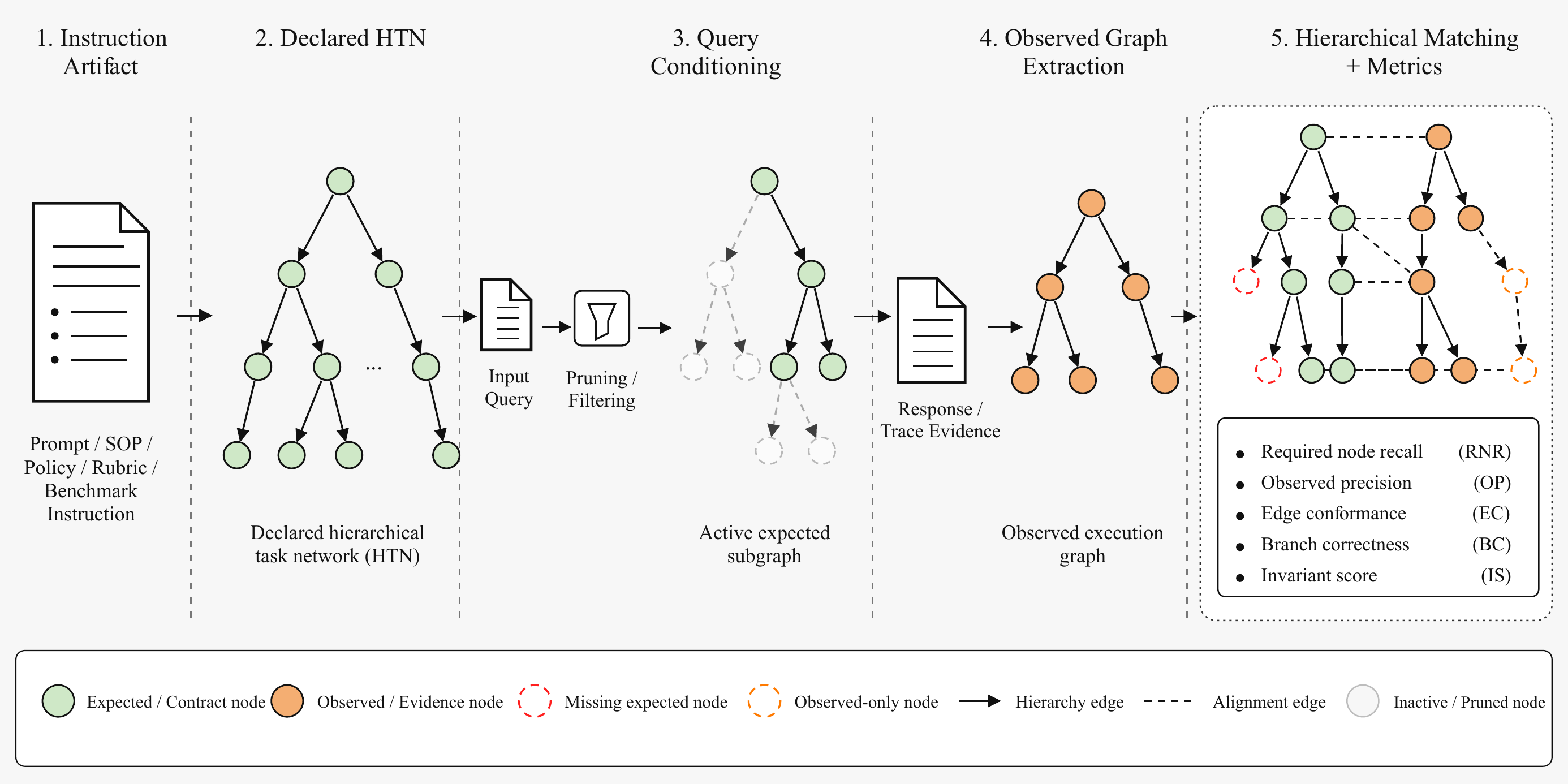}
  \caption{ContractEval pipeline. The key evaluation object is the query-active expected graph: only after the instruction artifact is conditioned on the user query can observed response or trace evidence be matched against the obligations it was supposed to support.}
  \label{fig:contracteval_pipeline}
\end{figure*}

\section{Introduction}

Modern LLM systems are increasingly asked not only to answer questions, but to
carry out procedures. A prompt, policy, rubric, or standard operating procedure
may require the system to check evidence, satisfy preconditions, choose the
right branch, respect dependencies, and preserve invariants such as privacy,
uncertainty, formatting, or tone. In such settings, the final answer is only
part of the behavior being evaluated. A response can be useful on the surface
while skipping the required verification, taking the wrong branch, reversing a
dependency, adding an unsupported intermediate action, or violating an invariant
that made the answer justified.

The difficulty is that these failures are often hidden behind plausible
outputs. A customer-service assistant may give a reasonable refund answer while
never checking eligibility; a medical or legal assistant may state a plausible
conclusion while omitting a required source check; an agent trace may contain
many tool calls while bypassing a prerequisite step. Output-only benchmarks and
holistic LLM judges can accept such cases because the answer reads well.
Trace-aware judges see more activity, but activity is not a denominator: the
trace alone does not say which obligations should have applied to this query,
which branch was active, or which observed actions count as satisfying the
procedure.

This turns procedural evaluation into a different question from answer grading:
not only whether the response is acceptable, but whether it was supported by the
required process. Without explicit active obligations, evaluators infer
that process from the same response or trace they are judging.

We introduce \textsc{ContractEval}, a framework for evaluating procedural
conformance by making this denominator explicit. ContractEval treats an
instruction artifact as a semantic contract, conditions it on the input query,
and compares the resulting active obligations with evidence extracted from the
response or trace. The central idea is falsifiable: a model should be evaluated
against the obligations that were active for this query, so failures can be
traced to missing or violated obligations rather than opaque judge preferences.

This does not require assuming that every valid behavior follows a single
surface trajectory. The query-conditioned expected graph is a partial-order
representation of semantic obligations: inactive branches are excluded,
required dependencies are retained, and observed behavior can match an
obligation at the appropriate level of abstraction. ContractEval is therefore
not a claim that agents must emit identical traces; it asks whether the trace or
response contains evidence that the required obligations were satisfied.

We test this idea with a controlled validity suite designed around the hidden
failure mode itself. From 10 audited procedural contracts and 200 clean
query-conditioned executions, we create paired cases in which the contract,
query, and base evidence remain fixed while one known obligation is omitted,
misbranched, reordered, supplemented with an extra action, or violated through
an invariant or output contract. This setup asks a simple question: when the
answer or trace still looks plausible, can an evaluator identify the failed
obligation? Across four judge models, output-only LLM judges achieve 0.495
average detection and 0.400 localization, and trace-aware judges reach 0.570
detection and 0.468 localization; the strongest trace-aware judge still misses
29.0\% of perturbed cases. Under audited expected and observed graphs,
ContractEval detects and localizes all controlled perturbations as a
construct-validity check, while LLM-backed observed extraction preserves much
of the signal but exposes extractor calibration as the scaling bottleneck.

Our contributions are: (1) a formulation of procedural conformance as matching
query-active obligations against observed behavior; (2) a query-conditioning
procedure that evaluates only obligations activated by the query while
preserving required dependencies; (3) a controlled perturbation benchmark for
testing hidden structural failures; (4) an empirical comparison of output-only,
trace-aware, expected-graph, oracle, and LLM-backed evaluator regimes; and (5)
a component study separating declared contract compilation, expected-subgraph
extraction, and observed-graph extraction. We will release the framework,
audited contracts, prompts, and benchmark artifacts to support reproducible
research on semantic conformance evaluation.

\section{Related Work}

\noindent\textbf{Instruction following and SOP evaluation.}\quad Instruction-following benchmarks have moved beyond single-answer accuracy by
testing whether models satisfy explicit constraints. IFEval focuses on
verifiable instruction constraints \citep{zhou2023instructionfollowingevaluationlargelanguage}; FollowBench,
InfoBench, and ComplexBench study increasingly complex multi-level instruction
following \citep{jiang-etal-2024-followbench,qin-etal-2024-infobench,NEURIPS2024_f8c24b08}; and
SOPBench and LongProc extend evaluation to longer procedural artifacts
\citep{nandi2026sopbenchcomplexindustrialsops,ye2025longproc}. SAGE further represents service SOPs as
dialogue graphs for logical compliance and path coverage \citep{shi2026sageserviceagentgraphguided}.
These benchmarks establish that instruction adherence is more structured than
final-answer correctness. ContractEval asks a different question: given a
particular query and an observed response or trace, which obligations from the
artifact were active, and did the observed behavior satisfy them? This shifts the
evaluation object from output adherence or path coverage to matching active
expected obligations against evidence of execution.

\noindent\textbf{LLM-as-a-judge and evaluator reliability.}\quad LLM-based evaluation is widely used for open-ended generation because it can
apply flexible rubrics and handle semantic variation. G-Eval, MT-Bench, Chatbot
Arena, and Prometheus-style evaluators exemplify this paradigm
\citep{liu-etal-2023-g,NEURIPS2023_91f18a12,ICLR2024_80348535,kim-etal-2024-prometheus}, while
surveys document both its usefulness and its sensitivity to prompts,
presentation, and rubric design \citep{gu2025surveyllmasajudge}. ContractEval is
motivated by a setting where holistic judgment is underspecified: a judge may
see a plausible final answer, or even a trace, without an explicit denominator
for what procedure should have applied. We therefore use LLMs, when needed, as
extractors or semantic verifiers inside a structured expected-vs-observed
protocol rather than as the sole source of the final conformance judgment.

\noindent\textbf{Agent traces, tools, and process evidence.}\quad Agent and tool-use benchmarks evaluate whether models can act through APIs,
browse, solve software tasks, or complete simulated tasks
\citep{yao2023react,NEURIPS2023_d842425e,ICLR2024_28e50ee5,liu2024agentbench,
zhou2024webarena,NEURIPS2024_5d413e48,jimenez2024swebench,ICLR2025_1b126cc3,
lu-etal-2025-toolsandbox,pmlr-v267-patil25a}. Recent work also evaluates traces directly,
including issue localization, goal-plan-action alignment, and trace-grounded
compliance \citep{deshpande2025trailtracereasoningagentic,jia2026agentsgpaframeworkevaluating,atf2025scenariobenchtracegroundedcomplianceevaluation}.
These settings make process evidence available, but evidence is not the same as
a protocol. A trace can be long, plausible, or tool-rich while still omitting a
required verification or following an inactive branch. ContractEval treats traces
as evidence for an observed graph and evaluates them only after conditioning the
artifact into query-active obligations; Table~\ref{tab:evaluator_gap}
summarizes the gap.

\begin{table}[t]
\centering
\small
\setlength{\tabcolsep}{3pt}
\renewcommand{\arraystretch}{1.06}
\begin{tabular}{@{}>{\raggedright\arraybackslash}p{0.31\columnwidth}>{\raggedright\arraybackslash}p{0.61\columnwidth}@{}}
\toprule
Evaluation family & ContractEval distinction \\
\midrule
Output-only judges & Check active obligations against evidence, not only final-answer plausibility. \\
Constraint benchmarks & Represent dependencies, branches, ordering, and invariants as query-conditioned obligations. \\
SOP/path benchmarks & Match expected obligations to observed behavior, not only path coverage. \\
Trace-aware judges & Expose the active denominator before judging trace evidence. \\
\bottomrule
\end{tabular}
\caption{How ContractEval differs from nearby evaluation families.}
\label{tab:evaluator_gap}
\end{table}

\noindent\textbf{Conformance checking, HTNs, and graph matching.}\quad ContractEval draws on ideas from process mining and conformance checking, where
event logs are compared with process models \citep{van_der_Aalst_2016,
Carmona_2018}, from hierarchical task networks
\citep{DBLP:conf/aaai/ErolHN94}, and from assignment-based graph matching
\citep{Kuhn_1955,Munkres_1957,Riesen_2009}. The
technical setting differs in two ways. First, the process model is not a
hand-authored formal workflow but a natural-language instruction artifact that
must be represented as semantic obligations. Second, the observed execution may
be implicit in a natural-language response or noisy agent trace. ContractEval
adapts conformance-style reasoning to this setting by constructing a
query-conditioned expected graph and matching it against evidence-grounded
observed behavior.

\section{Problem Setup}
\label{sec:problem_setup}

The preceding sections frame procedural evaluation as a question about
legitimacy: did the behavior satisfy the obligations that made the answer
warranted? We formalize this as conformance evaluation for a response or trace
conditioned on a natural-language instruction artifact. The evaluator receives
an artifact \(A\), a query \(q\), and observed behavior \(O\), and must determine
whether \(O\) satisfies the obligations that \(A\) makes active for \(q\).
Plausibility is not enough: a fluent or even factually correct conclusion can
still fail if it omits a required check, follows the wrong branch, or violates
an invariant along the way.

\noindent\textbf{Instruction artifacts.}\quad
An artifact \(A\) may be a prompt, SOP, policy, rubric, benchmark instruction,
or workflow description. It can specify actions, preconditions, decisions,
branch conditions, refusals, output schemas, and global constraints. Some
obligations are naturally graph-structured, such as ``verify eligibility before
approval.'' Others are response-level invariants, such as privacy, uncertainty,
formatting, non-fabrication, or tone. ContractEval keeps both kinds of
obligations in the evaluation target, but represents their structure
differently.

\noindent\textbf{Declared contract.}\quad
We represent the static obligations in \(A\) as a declared hierarchical task
network
\[
G_D=(V_D,E_D,H_D,I_D),
\]
where \(V_D\) contains semantic obligations, \(E_D\) contains directed
dependency edges, \(H_D\) is a parent-child hierarchy, and \(I_D\) is a set of
global invariants. Each node records a semantic type, description,
requiredness, and source evidence span. The hierarchy lets an artifact express
coarse tasks and finer substeps without forcing every evaluation to operate at
one fixed granularity.

\noindent\textbf{Query-conditioned denominator.}\quad
Only part of \(G_D\) is relevant for a particular query. ContractEval therefore
induces an expected graph
\[
G_E(q)=(V_E,E_E,H_E,I_E),
\]
by retaining active obligations, their prerequisites, their necessary
hierarchical context, and applicable invariants, while pruning inactive
branches. The denominator for conformance is \(G_E(q)\), not the full declared
contract. Crucially, \(G_E(q)\) is not a single prescribed execution trace. It
is a partial order over the semantic obligations that are normative for this
query. A valid system may combine adjacent steps, use different wording, or
provide richer evidence, provided that the required obligations and dependencies
are satisfied. Explicitly alternative procedures can be represented as branches
or equivalent obligation sets before matching. For example, if a refund SOP
requires eligibility verification before approval, \(G_E(q)\) requires evidence
of that check before the approval branch, not a particular wording.

\noindent\textbf{Observed behavior.}\quad
For the observed response or trace \(O\), we construct an observed execution
graph
\[
G_O=(V_O,E_O,H_O),
\]
whose nodes are evidence-grounded behaviors and whose edges capture observed
ordering or dependency relations. In the oracle experiments, \(G_O\) is derived
from audited gold executions and controlled perturbations. In scalable
experiments, it is extracted from response or trace text.

\noindent\textbf{Conformance objective.}\quad
Evaluation then asks whether \(G_O\) can be aligned to \(G_E(q)\). An alignment
\(M\subseteq V_E\times V_O\) maps expected obligations to observed evidence,
possibly across nearby levels of the HTN hierarchy. Unmatched required expected
nodes correspond to omissions; unmatched observed nodes correspond to extra
actions; violated expected edges indicate dependency or ordering failures; and
incorrect branch choices or invariant violations are scored separately. The
next section describes how ContractEval constructs these objects and converts
the alignment into diagnostic metrics.

\section{ContractEval Method}
\label{sec:method}

ContractEval operationalizes the problem setup as a staged evaluation pipeline
(Figure~\ref{fig:contracteval_pipeline}). The design separates questions that
holistic judging often collapses: what obligations the artifact declares, which
are active for this query, and what evidence the observed behavior provides.
Only then does ContractEval match expected and observed graphs and score
conformance failures, making representation, conditioning, extraction, and
matching quality separately reportable.

\noindent\textbf{1. Static contract compilation.}\quad
The compiler parses an instruction artifact into a declared HTN. Nodes are typed
as intents, actions, decisions, branch conditions, output requirements,
refusals, or validations. Edges encode semantic dependencies such as
\textit{requires}, \textit{enables}, \textit{branches\_to}, and
\textit{produces}; hierarchy edges encode decomposition. Each node stores a
short label, a natural-language description, requiredness, source evidence, and
metadata used by later conditioning. Global invariants are stored as separate
constraints because they usually apply to the whole response rather than to a
single ordered step.

\noindent\textbf{2. Query conditioning.}\quad
The conditioner maps a query \(q\) to the subset of the declared contract that
is active for that query. It selects the relevant branch decisions and their
outcomes, retains required prerequisites and ancestors, includes necessary
descendants for active composite tasks, and carries over applicable invariants.
Inactive branches are not counted as missed obligations. This step supplies the
expected graph \(G_E(q)\) used by all graph-topological metrics.

\noindent\textbf{3. Observed graph extraction.}\quad
ContractEval then constructs \(G_O\) from the available evidence. The evidence
may be a final response, a reasoning or tool trace, or a controlled execution
record. In oracle experiments, \(G_O\) is derived from audited gold executions
and injected perturbations. In scalable experiments, an LLM extractor emits a
JSON graph with observed-node descriptions, ordering edges, and evidence spans.
The reported extractor is blind to expected-node identifiers: it sees the
response or trace evidence, but not the answer key. Nodes without cited evidence,
or with cited evidence absent from the source text, are discarded. This design
makes extraction errors visible as missing observed evidence rather than allowing
the extractor to silently satisfy expected obligations.

\noindent\textbf{4. Hierarchical matching.}\quad
The matcher aligns expected obligations to observed evidence. Candidate matches
are generated between \(V_E\) and \(V_O\), scored using node labels,
descriptions, semantic types, and evidence spans, filtered by a fixed threshold,
and solved as a maximum-weight bipartite assignment
\citep{Kuhn_1955,Munkres_1957}. We use a fixed token-overlap
similarity in the reported LLM-backed runs and do not tune thresholds per model
or perturbation family. The hierarchy handles granularity mismatch: an observed
step may satisfy a leaf obligation, or may support an abstract parent when the
declared decomposition permits that level of evidence. Conversely, mandatory
children remain failures when the artifact requires them separately. The fixed
matching configuration is given in Appendix~\ref{sec:appendix_audit_matching}.

\noindent\textbf{Oracle and scalable regimes.}\quad
The same metric code is used in two regimes. In the oracle regime, declared,
expected, and observed graphs are audited or programmatically derived from gold
contracts, so matches can use gold semantic roles. This isolates representation
and metric validity: if a known omission, branch error, edge violation,
invariant breach, or output-contract failure is injected, the scores should move
in the intended dimension. In the scalable regime, declared and expected graphs
remain audited, but \(G_O\) is extracted from text and matched semantically.
This regime tests whether the signal survives realistic extraction noise; it is
not presented as fully automated compliance certification.

\noindent\textbf{Metrics.}\quad Let \(V_E^{req}\subseteq V_E\) be required expected nodes, and let
\(\mathrm{dom}(M)\) and \(\mathrm{rng}(M)\) be the mapped expected and observed
nodes. Required node recall is
\[
\mathrm{NodeRecall}=\frac{|V_E^{req}\cap \mathrm{dom}(M)|}{|V_E^{req}|}.
\]
Observed precision is
\[
\mathrm{ObservedPrecision}=\frac{|\mathrm{rng}(M)|}{|V_O|}.
\]
For edge conformance, let \(S_E\) be expected edges whose endpoints are mapped
and whose mapped observations satisfy the corresponding dependency; then
\(\mathrm{EdgeConf}=|S_E|/|E_E|\). Branch correctness averages whether observed
branch decisions match active expected decisions. Invariant score averages
applicable non-topological invariant scores \(s_i\in[0,1]\). In the oracle
perturbation setting, these scores come from audited invariant labels and the
injected breach type; in LLM-backed settings, a fixed invariant-judge prompt
scores the declared invariant against response or trace evidence. Output validity
is reported separately for strict final-output schemas. Together these dimensions
localize failures rather than collapsing them into one opaque scalar.

\section{Controlled Perturbation Evaluation Protocol}
\label{sec:protocol}

\begin{figure*}[t]
  \centering
  \includegraphics[width=0.82\linewidth]{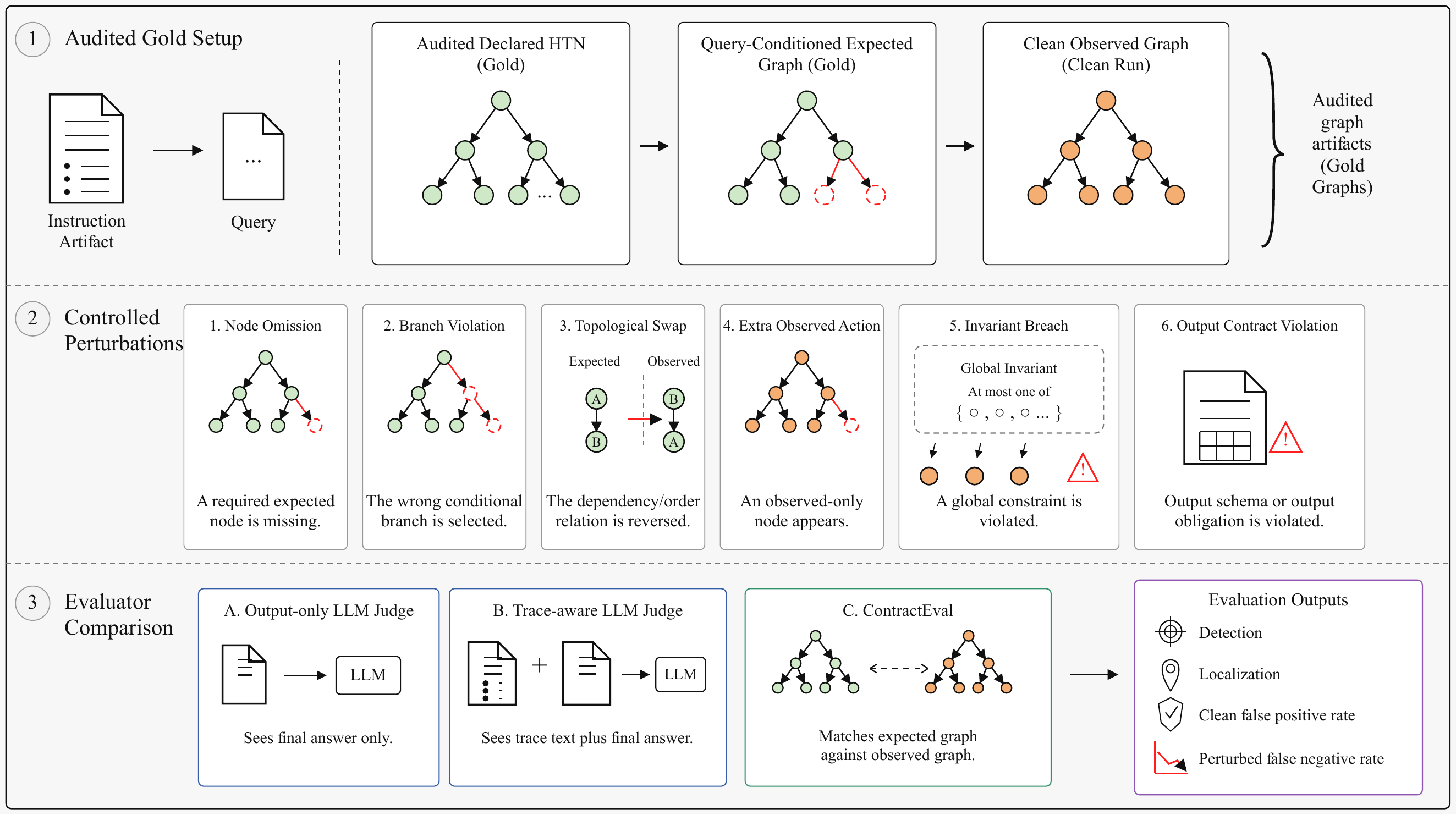}
  \caption{Audit and evaluation protocol. By holding the contract, query, and base evidence fixed while injecting one hidden procedural failure, the protocol tests whether an evaluator notices the failed obligation rather than merely accepting a plausible response or trace.}
  \label{fig:perturbation_protocol}
\end{figure*}

The perturbation protocol tests evaluator validity under known ground truth.
Human preference judgments do not identify which procedural obligation was
violated, while real agent telemetry entangles evaluator quality with tool
failures, logging choices, mocks, and prompt sensitivity. We therefore use
audited executions as clean substrates and inject one controlled structural
failure at a time, asking whether an evaluator detects and localizes that
failure (Figure~\ref{fig:perturbation_protocol}).

Each case starts from a human-audited declared HTN, a query-conditioned expected
graph, and a clean observed execution. We then generate six perturbation
families plus a clean control: \textit{node omission}, \textit{branch
violation}, \textit{topological swap}, \textit{extra observed action},
\textit{invariant breach}, \textit{output contract violation}, and
\textit{clean}. The families are chosen to target different parts of the
conformance definition. Omissions should reduce required-node recall and often
edge conformance; swaps should reduce ordering or dependency conformance; extra
actions should reduce observed precision; branch violations should reduce branch
correctness; invariant and output breaches should depress their corresponding
non-topological scores. The key stress test is hidden procedural failure:
several perturbations preserve a plausible final answer or trace while removing
evidence for a mandatory obligation. A compact example appears in
Appendix~\ref{sec:appendix_example}.

We evaluate four families of evaluators: output-only LLM judges, trace-aware LLM
judges, gold-graph ContractEval with audited declared/expected/observed graphs,
and ContractEval with audited declared/expected graphs but LLM-extracted
observed graphs. Detection is binary: clean controls should be predicted clean, and
perturbed cases should be predicted non-clean. Localization is stricter: the
predicted failure family must match the injected family, so detecting a problem
with the wrong label is not counted as localized. We report detection,
balanced detection accuracy, localization, clean false positive rate, and
perturbed false negative rate. The protocol is a construct-validity test: it
asks whether a diagnostic evaluator responds correctly when the missing, extra,
misordered, wrong-branch, invariant, or output-contract failure is known.

For ContractEval, the predicted family is obtained from the lowest affected
metric dimension after matching; if no dimension falls below its pre-specified
clean threshold, the case is predicted clean. The mapping is fixed before
evaluation: unmatched required nodes predict omissions, violated dependencies
predict swaps, unmatched observed nodes predict extras, and branch, invariant,
and output failures are assigned by their corresponding checks. No threshold or
tie order is tuned per model or perturbation family. Appendix~\ref{sec:appendix_audit_matching}
gives the full decision rule.

\section{Experimental Setup}
\label{sec:experimental_setup}

We evaluate ContractEval as an evaluation method, not as a leaderboard of task
solvers. The benchmark contains 10 SOP-style contracts, 20 queries per
contract, 200 clean query-conditioned executions, and 1,400 total cases after
adding six perturbation families to each clean case. The contracts are
intentionally deep rather than broad: declared HTNs contain 11.4 nodes and 11.1
edges per contract on average, with ranges up to 22 nodes and 27 edges, while
expected graphs contain 10.2 active nodes and 9.2 edges per query, with 3.6
global invariants per contract on average. The benchmark is
therefore a depth-oriented validity suite for hierarchical, dependency-rich
contracts; it is not intended to estimate average instruction-following
performance across domains.

For each contract, we maintain audited source instructions, declared HTNs,
query-conditioned expected graphs, and clean observed executions.
Perturbations are paired hidden-failure tests: clean and perturbed cases share
the same contract, query, and base evidence but differ in one known procedural
failure. We evaluate GPT-4o, GPT-5.4, Qwen3-32B, and Qwen2.5-7B. These results
should be read as evaluator and extractor behavior under a fixed validity suite,
not as a public ranking of the underlying models.

\noindent\textbf{Graph auditing.}\quad
The audited graphs provide the trusted substrate for the controlled setting.
Each declared HTN node is tied to a source evidence span, assigned a semantic
type, and checked for requiredness, parent-child hierarchy, and dependency
edges. Query-conditioned expected graphs are audited to ensure that inactive
branches are pruned while active prerequisites, ancestors, descendants, and
applicable invariants are retained. The audit scripts report zero schema errors
or warnings over the 10 contracts and validate evidence coverage for all
declared nodes. The primary construction is author-audited rather than a full
inter-annotator study. As a sanity check, a separate audit pass over two
contracts and 10 expected subgraphs found no path-changing disagreements
(Appendix~\ref{sec:appendix_audit_matching}). The pass checked whether declared
nodes were supported by source spans, whether dependency edges followed from
the SOP text, and whether query-conditioned graphs retained the active path.
Disagreements were recorded as granularity notes unless they changed the
required execution path. We still treat annotation cost, granularity
sensitivity, and full-corpus agreement as limitations. We use the audited
graphs as a construct-validity substrate, not as evidence that graph annotation
is cheap or uniquely determined.
Appendix~\ref{sec:appendix_audit_matching} details the audit checks, and
Appendix~\ref{sec:appendix_tables} reports domains and graph statistics.

\noindent\textbf{Evaluator modes.}\quad
We compare evaluator modes rather than official benchmark leaderboards because
existing instruction-following datasets do not define an evaluator for paired
hidden-failure traces. Output-only LLM judges see the instruction context,
query, and final response. Trace-aware judges additionally see execution or
trace text and, in the strongest prompt, a label-only expected-procedure
checklist. They do not receive declared HTN edges, query-conditioned graph
topology, an observed graph, or a matching result. All LLM judges return a
structured verdict with a clean/non-clean decision, failure family, and
rationale; metrics are computed from the parsed failure family rather than from
free-text rationales. Structured expected-graph LLM baselines receive the
serialized query-conditioned expected graph and the response or trace, but still
produce a direct judgment without deterministic graph matching. Oracle
Gold-graph ContractEval uses audited declared graphs, audited expected graphs,
and gold-grounded observed graphs. LLM-backed ContractEval keeps declared and
expected graphs audited but replaces the oracle observed graph with an
LLM-extracted observed graph.

\noindent\textbf{Component extraction.}\quad
We separately evaluate graph extraction against gold artifacts: declared HTN
compilation from the full instruction artifact, expected-graph extraction for a
query, and observed-graph extraction from response or trace text. Declared HTN
extraction is the hardest setting because the model must recover the full
static contract, including inactive branches. Expected-graph extraction is
easier because the query identifies the active obligations. Observed-graph
extraction is evaluated through downstream conformance, where hallucinated steps
become observed-only actions and missed evidence becomes unmatched obligations.
These component metrics are not conformance scores; they identify which parts
of the pipeline can be automated reliably and which should remain audited.
Because the perturbation suite contains 200 clean and 1,200 perturbed cases, we
report balanced detection accuracy alongside raw detection accuracy.

\noindent\textbf{Matching implementation.}\quad In the reported LLM-backed observed-extractor runs, candidate observed-to-expected
pairs are scored with the fixed token-overlap similarity used by our alignment
engine: normalized non-stopword tokens are compared with a weighted combination
of Jaccard overlap and containment. Pairs below the pre-specified 0.35
threshold are discarded, and the remaining one-to-one assignment is solved by a
maximum-weight bipartite assignment with a fixed deterministic solver. The
solver objective and threshold are fixed before evaluation; no model-specific
or perturbation-specific tuning is applied. The oracle setting instead uses
gold node identities, isolating metric behavior from extraction and matching
noise.

\section{Results}
\label{sec:results}

\subsection{LLM Judges Under-Detect Structural Failures}

Direct LLM judging is not sufficient for structural conformance evaluation
(full per-model results in Appendix~\ref{sec:appendix_tables}). Output-only
judges detect roughly half of the controlled failures on average. Trace text
helps, especially for the strongest model, but the trace-aware average remains
well below diagnostic use; GPT-5.4, the strongest trace-aware judge, still
misses 29.0\% of perturbed cases. These hidden failures show that evidence
alone does not define the expected execution structure against which it should
be checked.

\subsection{Gold-Graph ContractEval Checks Metric Validity}

\begin{table}[t]
\centering
\scriptsize
\setlength{\tabcolsep}{2pt}
\begin{tabular}{@{}lrrrrr@{}}
\toprule
Evaluator & Det. & Bal. & Loc. & FPR & FNR \\
\midrule
Output avg. & 0.495 & 0.695 & 0.400 & 0.026 & 0.584 \\
Trace avg. & 0.570 & 0.729 & 0.468 & 0.049 & 0.494 \\
Trace GPT-5.4 & 0.751 & 0.855 & 0.749 & 0.000 & 0.290 \\
ExpGraph GPT-4o & 0.676 & 0.802 & 0.669 & 0.020 & 0.375 \\
ExpGraph GPT-5.4 & 0.751 & 0.855 & 0.751 & 0.000 & 0.290 \\
ContractEval (gold graphs) & 1.000 & 1.000 & 1.000 & 0.000 & 0.000 \\
\bottomrule
\end{tabular}
\caption{Evaluator-family comparison. Gold-graph ContractEval measures metric validity under audited expected/observed graphs, not end-to-end automation. Bal. averages clean specificity and perturbed recall; FPR/FNR are clean false positives and perturbed false negatives.}
\label{tab:table:02:evaluator:family:comparison}
\end{table}

Table~\ref{tab:table:02:evaluator:family:comparison} compares evaluator
families. The expected-graph LLM judges are the strongest non-matching
baselines: they receive the query-conditioned expected graph and trace or final
answer, but still judge holistically. GPT-5.4 still marks 157/200 omissions and
185/200 swaps as clean; thus the gap to gold-graph ContractEval reflects
deterministic expected-observed matching, not merely graph exposure. This row is
a construct-validity check: with audited expected and observed graphs, the
metric responds correctly to each injected failure family. The LLM-backed
setting tests how much of that signal survives extraction.

\subsection{Metric Signatures Are Diagnostic}

Gold-graph scores move in the intended direction for each perturbation family
(Appendix~\ref{sec:appendix_tables}). Omissions reduce required-node recall and
edge conformance; swaps reduce ordering; extras reduce observed precision; and
branch, invariant, and output violations depress their corresponding scores.
The affected dimension identifies the kind of protocol failure, not just that a
case is worse.

\paragraph{LLM-backed extraction.}
With audited declared/expected graphs but LLM-extracted observed graphs,
ContractEval preserves much of the gold-graph signal: GPT-4o reaches 1.000/0.916
and GPT-5.4 reaches 0.982/0.903 detection/localization
(Appendix~\ref{sec:appendix_tables}). Residual errors concentrate in clean
false positives from extractor-added actions and in perturbation localization
misses, making observed-graph calibration the main scalability bottleneck.

\section{Discussion}
\label{sec:discussion}

\noindent\textbf{The unit of accountability matters.}\quad
Holistic judges do not merely lack trace evidence; they lack an explicit
account of what the trace is supposed to prove. ContractEval changes the unit of
evaluation from an answer-level verdict to evidence for query-active
obligations, making omissions, wrong branches, unsupported actions, and
invariant violations inspectable.

\noindent\textbf{Expected graphs are evaluation objects, not scripts.}\quad
This shift depends on reading the expected graph correctly. It is not a demand
for one surface trajectory, but an evaluation object: the obligations that
remain normative after conditioning on the query, including required checks,
dependencies, branch choices, and invariants. Hierarchical matching allows
semantically equivalent or differently worded behavior to satisfy an obligation
while preserving failures for missing prerequisites or invalid branch choices.

\noindent\textbf{The method changes what ``passing'' means.}\quad
Once the expected object is explicit, passing becomes an evidentiary claim. In
many LLM evaluations, a system passes if its final answer is acceptable or if a
judge can rationalize the answer from the trace. ContractEval instead asks
whether the observed behavior supports the obligations that were active for the
query. A model that reaches the right answer by skipping a required check has
not merely made a stylistic error; it has failed the contract that made the
answer legitimate.

\noindent\textbf{Automation should be staged.}\quad
The oracle setting validates the representation and metrics under audited
graphs; it does not show that graph extraction is solved. The practical value of
ContractEval is therefore a staged path from high-validity audits to scalable
benchmarking: audit declared contracts when correctness matters, automate
observed extraction where calibration is acceptable, and report component
quality alongside conformance scores.

\noindent\textbf{Evaluation should expose its denominator.}\quad
The broader implication is that process evaluations should report what
obligations they believed were active before judging whether behavior satisfied
them. Without this denominator, trace-aware judges can reward plausible evidence
from an inactive branch or overlook a missing prerequisite. ContractEval makes
that denominator inspectable, so disagreements about the expected procedure can
be audited rather than hidden inside a judge rationale.

\section{Conclusion}
\label{sec:conclusion}

ContractEval makes procedural conformance explicit by matching observed
behavior to query-active obligations. The controlled suite shows that this
exposes failures missed by holistic judges, while scalable use depends on
audited graph design and calibrated extraction. The broader lesson is that
process evaluation should ask whether the required procedure supports an
answer, not only whether the answer looks acceptable. As a diagnostic substrate,
ContractEval points toward naturalistic traces, independent annotation, and
calibrated extractors.

\clearpage
\section*{Limitations}

ContractEval is limited to text-centric SOP-style artifacts and controlled
response/trace evidence. The 10-contract, 1,400-case benchmark is
depth-oriented: contracts are hierarchical and dependency-rich, but omit
multimodal instructions, open-ended tool environments, and many organizational
policies. The results therefore support a controlled validity claim for
procedural conformance, not an estimate of average instruction-following
quality across deployment domains.

The perturbations isolate one failure at a time. Real failures may be compound,
ambiguous, repaired later, or intertwined with live tool failures, missing logs,
tool retries, and multi-turn recovery. Such settings may change what evidence
is available to the matcher and may require temporal or causal trace modeling
beyond the response/trace evidence used here. We therefore treat the benchmark
as a stress test for hidden procedural failures, not as evidence of deployment
robustness.

ContractEval also inherits expected-graph assumptions. Scores depend on graph
granularity, on which trajectories auditors deem equivalent, and on how partial
violations of invariants are calibrated. The second audit pass found no
path-changing disagreements in its sample, but full-corpus independent
agreement remains future work. In applications where the cost of a missed
obligation is high, declared contracts should remain audited and graph-design
choices should be reported with the conformance scores.

\section*{Ethical Considerations}

ContractEval is intended to improve transparency in evaluating whether systems follow declared instructions, but it should not be used as a standalone safety or compliance guarantee. Graph annotations and extractor prompts can encode author assumptions about which obligations matter, so released artifacts should include source evidence, schemas, prompts, and known limitations. When applied to real traces, evaluators may process sensitive user or operational data; users should minimize retained trace content, redact private information where possible, and report extractor uncertainty rather than present automated scores as definitive judgments. The controlled benchmark uses synthetic SOP-style artifacts and perturbations and is not a substitute for domain-specific risk assessment in high-stakes settings.

\bibliography{custom}

@String{june = "June"}

@misc{nandi2026sopbenchcomplexindustrialsops,
      title={SOP-Bench: Complex Industrial SOPs for Evaluating LLM Agents}, 
      author={Subhrangshu Nandi and Arghya Datta and Rohith Nama and Udita Patel and Nikhil Vichare and Indranil Bhattacharya and Prince Grover and Shivam Asija and Giuseppe Carenini and Wei Zhang and Arushi Gupta and Sreyoshi Bhaduri and Jing Xu and Huzefa Raja and Shayan Ray and Aaron Chan and Esther Xu Fei and Gaoyuan Du and Zuhaib Akhtar and Harshita Asnani and Weian Chan and Ming Xiong and Francesco Carbone and Jeetu Mirchandani},
      year={2026},
      eprint={2506.08119},
      archivePrefix={arXiv},
      primaryClass={cs.AI},
      url={https://arxiv.org/abs/2506.08119}, 
}

@misc{shi2026sageserviceagentgraphguided,
      title={SAGE: A Service Agent Graph-guided Evaluation Benchmark}, 
      author={Ling Shi and Yuqin Dai and Ziyin Wang and Ning Gao and Wei Zhang and Chaozheng Wang and Yujie Wang and Wei He and Jinpeng Wang and Deiyi Xiong},
      year={2026},
      eprint={2604.09285},
      archivePrefix={arXiv},
      primaryClass={cs.AI},
      url={https://arxiv.org/abs/2604.09285}, 
}

@misc{deshpande2025trailtracereasoningagentic,
      title={TRAIL: Trace Reasoning and Agentic Issue Localization}, 
      author={Darshan Deshpande and Varun Gangal and Hersh Mehta and Jitin Krishnan and Anand Kannappan and Rebecca Qian},
      year={2025},
      eprint={2505.08638},
      archivePrefix={arXiv},
      primaryClass={cs.AI},
      url={https://arxiv.org/abs/2505.08638}, 
}

@misc{jia2026agentsgpaframeworkevaluating,
      title={What Is Your Agent's GPA? A Framework for Evaluating Agent Goal-Plan-Action Alignment}, 
      author={Allison Sihan Jia and Daniel Huang and Nikhil Vytla and Seung Won Wilson Yoo and Nirvika Choudhury and Shayak Sen and John C. Mitchell and Anupam Datta},
      year={2026},
      eprint={2510.08847},
      archivePrefix={arXiv},
      primaryClass={cs.AI},
      url={https://arxiv.org/abs/2510.08847}, 
}

@misc{atf2025scenariobenchtracegroundedcomplianceevaluation,
      title={ScenarioBench: Trace-Grounded Compliance Evaluation for Text-to-SQL and RAG}, 
      author={Zahra Atf and Peter R Lewis},
      year={2025},
      eprint={2509.24212},
      archivePrefix={arXiv},
      primaryClass={cs.CL},
      url={https://arxiv.org/abs/2509.24212}, 
}

@inproceedings{
ye2025longproc,
title={LongProc: Benchmarking Long-Context Language Models on Long Procedural Generation},
author={Xi Ye and Fangcong Yin and Yinghui He and Joie Zhang and Howard Yen and Tianyu Gao and Greg Durrett and Danqi Chen},
booktitle={Second Conference on Language Modeling},
year={2025},
url={https://openreview.net/forum?id=ruWC5LIMSo}
}

@inproceedings{
yao2023react,
title={ReAct: Synergizing Reasoning and Acting in Language Models},
author={Shunyu Yao and Jeffrey Zhao and Dian Yu and Nan Du and Izhak Shafran and Karthik R Narasimhan and Yuan Cao},
booktitle={The Eleventh International Conference on Learning Representations },
year={2023},
url={https://openreview.net/forum?id=WE_vluYUL-X}
}

@inproceedings{NEURIPS2023_d842425e,
 author = {Schick, Timo and Dwivedi-Yu, Jane and Dessi, Roberto and Raileanu, Roberta and Lomeli, Maria and Hambro, Eric and Zettlemoyer, Luke and Cancedda, Nicola and Scialom, Thomas},
 booktitle = {Advances in Neural Information Processing Systems},
 editor = {A. Oh and T. Naumann and A. Globerson and K. Saenko and M. Hardt and S. Levine},
 pages = {68539--68551},
 publisher = {Curran Associates, Inc.},
 title = {Toolformer: Language Models Can Teach Themselves to Use Tools},
 url = {https://proceedings.neurips.cc/paper_files/paper/2023/file/d842425e4bf79ba039352da0f658a906-Paper-Conference.pdf},
 volume = {36},
 year = {2023}
}

@inproceedings{ICLR2024_28e50ee5,
 author = {Qin, Yujia and Liang, Shihao and Ye, Yining and Zhu, Kunlun and Yan, Lan and Lu, Yaxi and Lin, Yankai and Cong, Xin and Tang, Xiangru and Qian, Bill and Zhao, Sihan and Hong, Lauren and Tian, Runchu and Xie, Ruobing and Zhou, Jie and Gerstein, Mark and li, dahai and Liu, Zhiyuan and Sun, Maosong},
 booktitle = {International Conference on Learning Representations},
 editor = {B. Kim and Y. Yue and S. Chaudhuri and K. Fragkiadaki and M. Khan and Y. Sun},
 pages = {9695--9717},
 title = {ToolLLM: Facilitating Large Language Models to Master 16000+ Real-world APIs},
 url = {https://proceedings.iclr.cc/paper_files/paper/2024/file/28e50ee5b72e90b50e7196fde8ea260e-Paper-Conference.pdf},
 volume = {2024},
 year = {2024}
}

@inproceedings{
liu2024agentbench,
title={AgentBench: Evaluating {LLM}s as Agents},
author={Xiao Liu and Hao Yu and Hanchen Zhang and Yifan Xu and Xuanyu Lei and Hanyu Lai and Yu Gu and Hangliang Ding and Kaiwen Men and Kejuan Yang and Shudan Zhang and Xiang Deng and Aohan Zeng and Zhengxiao Du and Chenhui Zhang and Sheng Shen and Tianjun Zhang and Yu Su and Huan Sun and Minlie Huang and Yuxiao Dong and Jie Tang},
booktitle={The Twelfth International Conference on Learning Representations},
year={2024},
url={https://openreview.net/forum?id=zAdUB0aCTQ}
}

@inproceedings{
zhou2024webarena,
title={WebArena: A Realistic Web Environment for Building Autonomous Agents},
author={Shuyan Zhou and Frank F. Xu and Hao Zhu and Xuhui Zhou and Robert Lo and Abishek Sridhar and Xianyi Cheng and Tianyue Ou and Yonatan Bisk and Daniel Fried and Uri Alon and Graham Neubig},
booktitle={The Twelfth International Conference on Learning Representations},
year={2024},
url={https://openreview.net/forum?id=oKn9c6ytLx}
}

@inproceedings{NEURIPS2024_5d413e48,
 author = {Xie, Tianbao and Zhang, Danyang and Chen, Jixuan and Li, Xiaochuan and Zhao, Siheng and Cao, Ruisheng and Hua, Toh Jing and Cheng, Zhoujun and Shin, Dongchan and Lei, Fangyu and Liu, Yitao and Xu, Yiheng and Zhou, Shuyan and Savarese, Silvio and Xiong, Caiming and Zhong, Victor and Yu, Tao},
 booktitle = {Advances in Neural Information Processing Systems},
 doi = {10.52202/079017-1650},
 editor = {A. Globerson and L. Mackey and D. Belgrave and A. Fan and U. Paquet and J. Tomczak and C. Zhang},
 pages = {52040--52094},
 publisher = {Curran Associates, Inc.},
 title = {OSWorld: Benchmarking Multimodal Agents for Open-Ended Tasks in Real Computer Environments},
 url = {https://proceedings.neurips.cc/paper_files/paper/2024/file/5d413e48f84dc61244b6be550f1cd8f5-Paper-Datasets_and_Benchmarks_Track.pdf},
 volume = {37},
 year = {2024}
}

@inproceedings{
jimenez2024swebench,
title={{SWE}-bench: Can Language Models Resolve Real-world Github Issues?},
author={Carlos E Jimenez and John Yang and Alexander Wettig and Shunyu Yao and Kexin Pei and Ofir Press and Karthik R Narasimhan},
booktitle={The Twelfth International Conference on Learning Representations},
year={2024},
url={https://openreview.net/forum?id=VTF8yNQM66}
}

@inproceedings{ICLR2025_1b126cc3,
 author = {Yao, Shunyu and Shinn, Noah and Razavi, Pedram and Narasimhan, Karthik},
 booktitle = {International Conference on Learning Representations},
 editor = {Y. Yue and A. Garg and N. Peng and F. Sha and R. Yu},
 pages = {9965--10017},
 title = {{$\tau$}-bench: A Benchmark for \underline{T}ool-\underline{A}gent-\underline{U}ser Interaction in Real-World Domains},
 url = {https://proceedings.iclr.cc/paper_files/paper/2025/file/1b126cc38b8638e07bef37e7b2bb72bf-Paper-Conference.pdf},
 volume = {2025},
 year = {2025}
}

@inproceedings{lu-etal-2025-toolsandbox,
    title = "{T}ool{S}andbox: A Stateful, Conversational, Interactive Evaluation Benchmark for {LLM} Tool Use Capabilities",
    author = "Lu, Jiarui  and
      Holleis, Thomas  and
      Zhang, Yizhe  and
      Aumayer, Bernhard  and
      Nan, Feng  and
      Bai, Haoping  and
      Ma, Shuang  and
      Ma, Shen  and
      Li, Mengyu  and
      Yin, Guoli  and
      Wang, Zirui  and
      Pang, Ruoming",
    editor = "Chiruzzo, Luis  and
      Ritter, Alan  and
      Wang, Lu",
    booktitle = "Findings of the Association for Computational Linguistics: NAACL 2025",
    month = apr,
    year = "2025",
    address = "Albuquerque, New Mexico",
    publisher = "Association for Computational Linguistics",
    url = "https://aclanthology.org/2025.findings-naacl.65/",
    doi = "10.18653/v1/2025.findings-naacl.65",
    pages = "1160--1183",
    ISBN = "979-8-89176-195-7"
}

@InProceedings{pmlr-v267-patil25a,
  title = 	 {The Berkeley Function Calling Leaderboard ({BFCL}): From Tool Use to Agentic Evaluation of Large Language Models},
  author =       {Patil, Shishir G and Mao, Huanzhi and Yan, Fanjia and Ji, Charlie Cheng-Jie and Suresh, Vishnu and Stoica, Ion and Gonzalez, Joseph E.},
  booktitle = 	 {Proceedings of the 42nd International Conference on Machine Learning},
  pages = 	 {48371--48392},
  year = 	 {2025},
  editor = 	 {Singh, Aarti and Fazel, Maryam and Hsu, Daniel and Lacoste-Julien, Simon and Berkenkamp, Felix and Maharaj, Tegan and Wagstaff, Kiri and Zhu, Jerry},
  volume = 	 {267},
  series = 	 {Proceedings of Machine Learning Research},
  month = 	 {13--19 Jul},
  publisher =    {PMLR},
  url = 	 {https://proceedings.mlr.press/v267/patil25a.html}
}

@inproceedings{liu-etal-2023-g,
    title = "{G}-Eval: {NLG} Evaluation using Gpt-4 with Better Human Alignment",
    author = "Liu, Yang  and
      Iter, Dan  and
      Xu, Yichong  and
      Wang, Shuohang  and
      Xu, Ruochen  and
      Zhu, Chenguang",
    editor = "Bouamor, Houda  and
      Pino, Juan  and
      Bali, Kalika",
    booktitle = "Proceedings of the 2023 Conference on Empirical Methods in Natural Language Processing",
    month = dec,
    year = "2023",
    address = "Singapore",
    publisher = "Association for Computational Linguistics",
    url = "https://aclanthology.org/2023.emnlp-main.153/",
    doi = "10.18653/v1/2023.emnlp-main.153",
    pages = "2511--2522"
}

@inproceedings{NEURIPS2023_91f18a12,
 author = {Zheng, Lianmin and Chiang, Wei-Lin and Sheng, Ying and Zhuang, Siyuan and Wu, Zhanghao and Zhuang, Yonghao and Lin, Zi and Li, Zhuohan and Li, Dacheng and Xing, Eric and Zhang, Hao and Gonzalez, Joseph and Stoica, Ion},
 booktitle = {Advances in Neural Information Processing Systems},
 editor = {A. Oh and T. Naumann and A. Globerson and K. Saenko and M. Hardt and S. Levine},
 pages = {46595--46623},
 publisher = {Curran Associates, Inc.},
 title = {Judging LLM-as-a-Judge with MT-Bench and Chatbot Arena},
 url = {https://proceedings.neurips.cc/paper_files/paper/2023/file/91f18a1287b398d378ef22505bf41832-Paper-Datasets_and_Benchmarks.pdf},
 volume = {36},
 year = {2023}
}

@inproceedings{ICLR2024_80348535,
 author = {Kim, Seungone and Shin, Jay and cho, yejin and Jang, Joel and Longpre, Shayne and Lee, Hwaran and Yun, Sangdoo and Shin, Ryan, S and Kim, Sungdong and Thorne, James and Seo, Minjoon},
 booktitle = {International Conference on Learning Representations},
 editor = {B. Kim and Y. Yue and S. Chaudhuri and K. Fragkiadaki and M. Khan and Y. Sun},
 pages = {29927--29962},
 title = {Prometheus: Inducing Fine-Grained Evaluation Capability in Language Models},
 url = {https://proceedings.iclr.cc/paper_files/paper/2024/file/803485352e61e3ebf41221e4776c9fd4-Paper-Conference.pdf},
 volume = {2024},
 year = {2024}
}

@inproceedings{kim-etal-2024-prometheus,
    title = "Prometheus 2: An Open Source Language Model Specialized in Evaluating Other Language Models",
    author = "Kim, Seungone  and
      Suk, Juyoung  and
      Longpre, Shayne  and
      Lin, Bill Yuchen  and
      Shin, Jamin  and
      Welleck, Sean  and
      Neubig, Graham  and
      Lee, Moontae  and
      Lee, Kyungjae  and
      Seo, Minjoon",
    editor = "Al-Onaizan, Yaser  and
      Bansal, Mohit  and
      Chen, Yun-Nung",
    booktitle = "Proceedings of the 2024 Conference on Empirical Methods in Natural Language Processing",
    month = nov,
    year = "2024",
    address = "Miami, Florida, USA",
    publisher = "Association for Computational Linguistics",
    url = "https://aclanthology.org/2024.emnlp-main.248/",
    doi = "10.18653/v1/2024.emnlp-main.248",
    pages = "4334--4353"
}

@misc{gu2025surveyllmasajudge,
      title={A Survey on LLM-as-a-Judge}, 
      author={Jiawei Gu and Xuhui Jiang and Zhichao Shi and Hexiang Tan and Xuehao Zhai and Chengjin Xu and Wei Li and Yinghan Shen and Shengjie Ma and Honghao Liu and Saizhuo Wang and Kun Zhang and Yuanzhuo Wang and Wen Gao and Lionel Ni and Jian Guo},
      year={2025},
      eprint={2411.15594},
      archivePrefix={arXiv},
      primaryClass={cs.CL},
      url={https://arxiv.org/abs/2411.15594}, 
}

@misc{zhou2023instructionfollowingevaluationlargelanguage,
      title={Instruction-Following Evaluation for Large Language Models}, 
      author={Jeffrey Zhou and Tianjian Lu and Swaroop Mishra and Siddhartha Brahma and Sujoy Basu and Yi Luan and Denny Zhou and Le Hou},
      year={2023},
      eprint={2311.07911},
      archivePrefix={arXiv},
      primaryClass={cs.CL},
      url={https://arxiv.org/abs/2311.07911}, 
}

@inproceedings{jiang-etal-2024-followbench,
    title = "{F}ollow{B}ench: A Multi-level Fine-grained Constraints Following Benchmark for Large Language Models",
    author = "Jiang, Yuxin  and
      Wang, Yufei  and
      Zeng, Xingshan  and
      Zhong, Wanjun  and
      Li, Liangyou  and
      Mi, Fei  and
      Shang, Lifeng  and
      Jiang, Xin  and
      Liu, Qun  and
      Wang, Wei",
    editor = "Ku, Lun-Wei  and
      Martins, Andre  and
      Srikumar, Vivek",
    booktitle = "Proceedings of the 62nd Annual Meeting of the Association for Computational Linguistics (Volume 1: Long Papers)",
    month = aug,
    year = "2024",
    address = "Bangkok, Thailand",
    publisher = "Association for Computational Linguistics",
    url = "https://aclanthology.org/2024.acl-long.257/",
    doi = "10.18653/v1/2024.acl-long.257",
    pages = "4667--4688"
}

@inproceedings{qin-etal-2024-infobench,
    title = "{I}n{F}o{B}ench: Evaluating Instruction Following Ability in Large Language Models",
    author = "Qin, Yiwei  and
      Song, Kaiqiang  and
      Hu, Yebowen  and
      Yao, Wenlin  and
      Cho, Sangwoo  and
      Wang, Xiaoyang  and
      Wu, Xuansheng  and
      Liu, Fei  and
      Liu, Pengfei  and
      Yu, Dong",
    editor = "Ku, Lun-Wei  and
      Martins, Andre  and
      Srikumar, Vivek",
    booktitle = "Findings of the Association for Computational Linguistics: ACL 2024",
    month = aug,
    year = "2024",
    address = "Bangkok, Thailand",
    publisher = "Association for Computational Linguistics",
    url = "https://aclanthology.org/2024.findings-acl.772/",
    doi = "10.18653/v1/2024.findings-acl.772",
    pages = "13025--13048"
}

@inproceedings{NEURIPS2024_f8c24b08,
 author = {Wen, Bosi and Ke, Pei and Gu, Xiaotao and Wu, Lindong and Huang, Hao and Zhou, Jinfeng and Li, Wenchuang and Hu, Binxin and Gao, Wendy and Xu, Jiaxin and Liu, Yiming and Tang, Jie and Wang, Hongning and Huang, Minlie},
 booktitle = {Advances in Neural Information Processing Systems},
 doi = {10.52202/079017-4371},
 editor = {A. Globerson and L. Mackey and D. Belgrave and A. Fan and U. Paquet and J. Tomczak and C. Zhang},
 pages = {137610--137645},
 publisher = {Curran Associates, Inc.},
 title = {Benchmarking Complex Instruction-Following with Multiple Constraints Composition},
 url = {https://proceedings.neurips.cc/paper_files/paper/2024/file/f8c24b08b96a08ec7a7a975feea7777e-Paper-Datasets_and_Benchmarks_Track.pdf},
 volume = {37},
 year = {2024}
}

@article{Kuhn_1955, title={The Hungarian method for the assignment problem}, volume={2}, ISSN={1931-9193}, url={http://dx.doi.org/10.1002/nav.3800020109}, DOI={10.1002/nav.3800020109}, number={1-2}, journal={Naval Research Logistics Quarterly}, publisher={Wiley}, author={Kuhn, H. W.}, year={1955}, month=Mar, pages={83–97} }

@article{Munkres_1957, title={Algorithms for the Assignment and Transportation Problems}, volume={5}, ISSN={2168-3484}, url={http://dx.doi.org/10.1137/0105003}, DOI={10.1137/0105003}, number={1}, journal={Journal of the Society for Industrial and Applied Mathematics}, publisher={Society for Industrial & Applied Mathematics (SIAM)}, author={Munkres, James}, year={1957}, month=Mar, pages={32–38} }

@article{Riesen_2009, title={Approximate graph edit distance computation by means of bipartite graph matching}, volume={27}, ISSN={0262-8856}, url={http://dx.doi.org/10.1016/j.imavis.2008.04.004}, DOI={10.1016/j.imavis.2008.04.004}, number={7}, journal={Image and Vision Computing}, publisher={Elsevier BV}, author={Riesen, Kaspar and Bunke, Horst}, year={2009}, month=June, pages={950–959} }

@inproceedings{DBLP:conf/aaai/ErolHN94,
  author       = {Kutluhan Erol and
                  James A. Hendler and
                  Dana S. Nau},
  editor       = {Barbara Hayes{-}Roth and
                  Richard E. Korf},
  title        = {{HTN} Planning: Complexity and Expressivity},
  booktitle    = {Proceedings of the 12th National Conference on Artificial Intelligence,
                  Seattle, WA, USA, July 31 - August 4, 1994, Volume 2},
  pages        = {1123--1128},
  publisher    = {{AAAI} Press / The {MIT} Press},
  year         = {1994},
  url          = {http://www.aaai.org/Library/AAAI/1994/aaai94-173.php},
  bibsource    = {dblp computer science bibliography, https://dblp.org}
}

@book{van_der_Aalst_2016, title={Process Mining}, ISBN={9783662498514}, url={http://dx.doi.org/10.1007/978-3-662-49851-4}, DOI={10.1007/978-3-662-49851-4}, publisher={Springer Berlin Heidelberg}, author={van der Aalst, Wil}, year={2016} }

@book{Carmona_2018, title={Conformance Checking: Relating Processes and Models}, ISBN={9783319994147}, url={http://dx.doi.org/10.1007/978-3-319-99414-7}, DOI={10.1007/978-3-319-99414-7}, publisher={Springer International Publishing}, author={Carmona, Josep and van Dongen, Boudewijn and Solti, Andreas and Weidlich, Matthias}, year={2018} }
\clearpage
\appendix

% \raggedbottom
\section{Gold Graph Audit and Matching Details}
\label{sec:appendix_audit_matching}

\noindent\textbf{Audit protocol.}\quad Each SOP artifact is converted into a
declared HTN with semantic nodes, typed dependency edges, hierarchy links, and
global invariants. The audit checks that each node has a source evidence span,
that requiredness and node type are explicit, that branch conditions are
represented as decision obligations rather than hidden prose, and that global
invariants are not encoded as ordered graph steps. Query-conditioned expected
graphs are then checked against each task row to ensure that active branches,
prerequisites, ancestors, descendants, and applicable invariants are retained
while inactive branches are excluded. The released audit artifact records zero
schema errors or warnings for the 10 contracts and validates evidence spans for
all declared nodes. This audit is an author-audited construction rather than a
multi-annotator agreement study; we therefore do not use it as evidence that
graph annotation is trivial or subjective choices disappear.

\noindent\textbf{Second audit pass.}\quad After the primary author audit, a
separate audit pass reviewed two contracts (\textit{customer\_service} and
\textit{dangerous\_goods}), covering 35 declared nodes, 40 declared edges, and
10 sampled query-conditioned expected subgraphs. The pass found 35/35 declared
nodes and 40/40 declared edges to be supported by the SOP text and found no
path-changing disagreement in the sampled expected subgraphs. It recorded four
minor granularity notes: suspension/payment handling and final documentation in
\textit{customer\_service}, and per-component score validation and audit-log
outputs in \textit{dangerous\_goods}, are represented by coarser semantic nodes
rather than separate obligations. We report this as a sanity check, not a full
inter-annotator agreement study; full-corpus independent annotation and timing
remain future work.

\noindent\textbf{Observed matching configuration.}\quad Oracle ContractEval
uses gold node identities in the controlled perturbation suite. In LLM-backed
observed extraction, the extractor emits observed nodes with short descriptions
and evidence spans. ContractEval then computes a fixed token-similarity matrix
between observed and expected node descriptions. Tokens are lowercased, split on
punctuation and separators, normalized for a small set of aliases and suffixes,
and stripped of stopwords. Pair score is \(0.65\) times Jaccard overlap plus
\(0.35\) times containment over the smaller token set. Candidate pairs below
0.35 are rejected, and the remaining assignment is solved by maximum-weight
bipartite matching. Thresholds and scoring rules are fixed before evaluation; no
per-model post-hoc tuning is applied in the reported tables.

\noindent\textbf{Metric-vector decision rule.}\quad For ContractEval detection
and localization, the predicted family is obtained from the lowest affected
metric dimension after matching. Unmatched required expected nodes predict node
omission; unmatched observed nodes predict extra observed action; violated
dependencies between matched nodes predict topological swap; failed branch
decisions predict branch violation; and failed invariant or output checks
predict their corresponding labels. If no dimension falls below its
pre-specified clean threshold, the case is predicted clean. Ties are resolved by
a fixed priority order matching the injected-family taxonomy.

\section{Worked Example}
\label{sec:appendix_example}

\begin{figure*}[t]
\centering
\includegraphics[width=\textwidth]{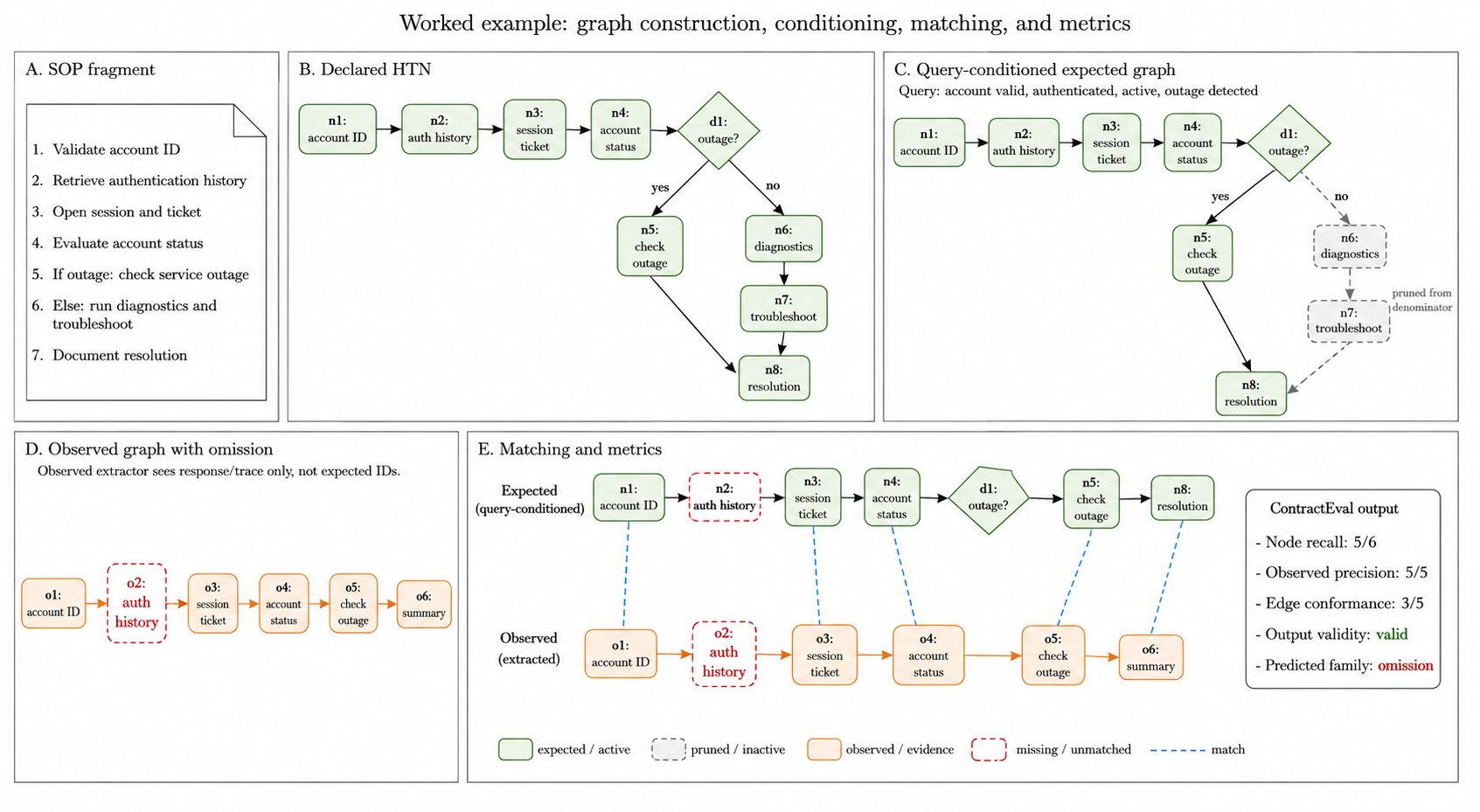}
\caption{Worked example showing the concrete graph objects compared by ContractEval: an SOP fragment, declared HTN, query-conditioned expected graph, observed graph, matching assignment, and resulting diagnostic metrics.}
\label{fig:worked_example_graphs}
\end{figure*}

Figure~\ref{fig:worked_example_graphs} shows the concrete objects in one
abbreviated customer-service case. The example is schematic; the released
artifact contains the full JSON graphs and evidence spans.

The key point is that the no-outage branch remains in the declared HTN but is
not part of the query-conditioned denominator. The observed omission leaves the
expected authentication node unmatched, which reduces node recall and invalidates
downstream dependency edges while leaving observed precision high.

\section{Benchmark and Component Tables}
\label{sec:appendix_tables}

\begin{table}[H]
\centering
\small
\begin{tabular}{lr}
\toprule
Quantity & Value \\
\midrule
SOP-style contracts & 10 \\
Queries / expected graphs & 200 \\
Clean executions & 200 \\
Perturbation categories incl. clean & 7 \\
Total evaluation cases & 1400 \\
Declared HTN nodes / contract & 11.4 [6 to 22] \\
Declared HTN edges / contract & 11.1 [5 to 27] \\
Branch nodes / contract & 0.6 [0 to 5] \\
Global invariants / contract & 3.6 [3 to 5] \\
Expected nodes / query & 10.2 [5 to 16] \\
Expected edges / query & 9.2 [4 to 15] \\
\bottomrule
\end{tabular}

\caption{Controlled perturbation benchmark summary. Ranges are reported as mean [min to max].}
\label{tab:table:00:dataset:summary}
\end{table}

\begin{table*}[t]
\centering
\scriptsize
\resizebox{0.65\textwidth}{!}{%
\begin{tabular}{lrrrr}
\toprule
Contract & Nodes & Edges & Branches & Inv. \\
\midrule
Aircraft inspection & 9 & 8 & 0 & 3 \\
Content flagging & 12 & 11 & 0 & 4 \\
Customer service & 22 & 27 & 5 & 4 \\
Dangerous goods & 13 & 13 & 1 & 4 \\
Know your business & 14 & 13 & 0 & 4 \\
Order fulfillment & 6 & 5 & 0 & 3 \\
Patient intake & 9 & 8 & 0 & 3 \\
Referral abuse detection v1 & 9 & 8 & 0 & 3 \\
Referral abuse detection v2 & 12 & 11 & 0 & 5 \\
Traffic spoofing detection & 8 & 7 & 0 & 3 \\
\bottomrule
\end{tabular}
}%
\caption{SOP contract domains and declared-HTN complexity. Each contract contributes 20 query-conditioned expected graphs.}
\label{tab:sop_domains}
\end{table*}

\begin{table*}[t]
\centering
\scriptsize
\resizebox{\textwidth}{!}{%
\begin{tabular}{lrrrrrr}
\toprule
Model & Node F1 & Node Exact & Edge F1 & Edge Exact & Branch F1 & Branch Exact \\
\midrule
GPT-4o & 0.977 & 0.890 & 0.970 & 0.890 & 0.800 & 0.800 \\
GPT-5.4 & 0.986 & 0.910 & 0.980 & 0.910 & 0.800 & 0.800 \\
Qwen3-32B & 0.983 & 0.895 & 0.979 & 0.925 & 0.750 & 0.750 \\
Qwen2.5-7B & 0.965 & 0.870 & 0.965 & 0.910 & 0.510 & 0.510 \\
\bottomrule
\end{tabular}
}%
\caption{Query-conditioned expected-subgraph extraction against gold expected graphs.}
\label{tab:table:05:expected:subgraph:extraction}
\end{table*}

\begin{table*}[t]
\centering
\scriptsize
\resizebox{\textwidth}{!}{%
\begin{tabular}{lrrrrrrr}
\toprule
Model & Node P & Node R & Edge P & Edge R & Hierarchy & Branch R & GED Sim. \\
\midrule
GPT-4o & 0.438 & 0.381 & 0.295 & 0.095 & 0.560 & 0.800 & 0.332 \\
GPT-5.4 & 0.322 & 0.552 & 0.225 & 0.312 & 0.200 & 0.880 & 0.409 \\
Qwen3-32B & 0.636 & 0.171 & 0.532 & 0.020 & 0.600 & 0.800 & 0.242 \\
Qwen2.5-7B & 0.310 & 0.218 & 0.196 & 0.062 & 0.700 & 0.800 & 0.280 \\
\bottomrule
\end{tabular}
}%
\caption{Declared StaticHTN extraction against human-audited gold contracts.}
\label{tab:table:06:static:htn:extraction}
\end{table*}

\begin{table*}[t]
\centering
\scriptsize
\resizebox{\textwidth}{!}{%
\begin{tabular}{lrrrrrr}
\toprule
Model & Static GED & Static Node R & Expected Node F1 & Expected Edge F1 & Obs.-backed Det. & Obs.-backed Loc. \\
\midrule
GPT-4o & 0.332 & 0.381 & 0.977 & 0.970 & 1.000 & 0.916 \\
GPT-5.4 & 0.409 & 0.552 & 0.986 & 0.980 & 0.982 & 0.903 \\
Qwen3-32B & 0.242 & 0.171 & 0.983 & 0.979 & 0.986 & 0.846 \\
Qwen2.5-7B & 0.280 & 0.218 & 0.965 & 0.965 & 0.944 & 0.660 \\
\bottomrule
\end{tabular}
}%
\caption{Compact component scaling summary separating static contract compilation, expected-subgraph extraction, and observed-graph-backed conformance.}
\label{tab:table:07:component:scaling:compact}
\end{table*}

\begin{table}[t]
\centering
\small
\setlength{\tabcolsep}{3pt}
\begin{tabular}{@{}lrrrr@{}}
\toprule
Metric & GPT-4o & GPT-5.4 & Qwen3 & Qwen2.5 \\
\midrule
Output det. & 0.501 & 0.571 & 0.490 & 0.419 \\
Output loc. & 0.471 & 0.525 & 0.278 & 0.326 \\
Trace det. & 0.506 & 0.751 & 0.606 & 0.415 \\
Trace loc. & 0.494 & 0.749 & 0.358 & 0.269 \\
\bottomrule
\end{tabular}
\caption{LLM judge detection and localization on the 1,400-case perturbation suite. Qwen3 denotes Qwen3-32B; Qwen2.5 denotes Qwen2.5-7B.}
\label{tab:table:01:llm:judge:baselines}
\end{table}

\begin{table}[t]
\centering
\small
\setlength{\tabcolsep}{3pt}
\begin{tabular}{@{}lll@{}}
\toprule
Perturb. & Metric(s) & Value \\
\midrule
Clean & none & all 1.000 \\
Node omission & Node recall; edge & 0.895; 0.762 \\
Branch viol. & Branch; output & 0.000; 0.000 \\
Topol. swap & Edge order & 0.881 \\
Extra action & Obs. precision & 0.905 \\
Invariant & Invariant & 0.000 \\
Output & Output & 0.000 \\
\bottomrule
\end{tabular}
\caption{Compact oracle metric sensitivity. Each row has 200 cases; unlisted metrics remain 1.000.}
\label{tab:table:03:oracle:metric:sensitivity}
\end{table}

\begin{table}[t]
\centering
\small
\setlength{\tabcolsep}{3pt}
\begin{tabular}{@{}lrrrr@{}}
\toprule
Metric & GPT-4o & GPT-5.4 & Qwen3 & Qwen2.5 \\
\midrule
Detection & 1.000 & 0.982 & 0.986 & 0.944 \\
Bal. acc. & 1.000 & 0.938 & 0.950 & 0.802 \\
Localization & 0.916 & 0.903 & 0.846 & 0.660 \\
Clean FPR & 0.000 & 0.125 & 0.100 & 0.395 \\
Perturbed FNR & 0.000 & 0.000 & 0.000 & 0.000 \\
\bottomrule
\end{tabular}
\caption{LLM-backed ContractEval on 1,400 cases; Qwen rows denote Qwen3-32B and Qwen2.5-7B.}
\label{tab:table:04:llm:backed:contracteval}
\end{table}

\section{Per-Family Localization}
\label{sec:appendix_per_family}

The per-family view shows that free-form judges do not fail uniformly. Average
output-only judges are strong on clean cases and branch failures that alter the
final answer, but cannot reliably identify omitted steps, topological swaps, or
extra trace actions. Trace-aware judging improves extra-action recognition, and
the best trace-aware GPT-5.4 run is strong on branch, invariant, output, and
extra-action failures, but still has low recall for omissions and ordering
swaps. LLM-backed ContractEval shifts the error profile: most perturbation
families remain high, while remaining failures reflect observed-extractor
calibration, especially whether clean traces are over-segmented or extra actions
are conservatively emitted.

\begin{table*}[t]
\centering
\scriptsize
\resizebox{\textwidth}{!}{%
\begin{tabular}{lccccccc}
\toprule
Evaluator & Clean & Node omit. & Branch & Swap & Extra & Invariant & Output \\
\midrule
Avg. output-only LLM & 0.974 & 0.000 & 0.767 & 0.000 & 0.000 & 0.611 & 0.449 \\
Avg. trace-aware LLM & 0.951 & 0.129 & 0.848 & 0.018 & 0.401 & 0.519 & 0.407 \\
Best trace-aware LLM (GPT-5.4) & 1.000 & 0.250 & 1.000 & 0.070 & 0.965 & 0.985 & 0.975 \\
Oracle ContractEval & 1.000 & 1.000 & 1.000 & 1.000 & 1.000 & 1.000 & 1.000 \\
ContractEval + GPT-4o observed extractor & 1.000 & 1.000 & 1.000 & 1.000 & 0.415 & 1.000 & 1.000 \\
ContractEval + GPT-5.4 observed extractor & 0.875 & 1.000 & 0.900 & 0.915 & 0.855 & 0.910 & 0.865 \\
\bottomrule
\end{tabular}
}%
\caption{Per-family exact-label recall. Clean is exact clean classification; other columns require predicting the injected perturbation family, not merely detecting that a case is perturbed.}
\label{tab:per_family_localization}
\end{table*}

\section{Confusion Matrices}
\label{sec:appendix_confusion}

The confusion matrices make the localization failures more explicit. The
strongest trace-aware and expected-graph LLM judges still confuse omissions and
ordering failures with clean executions in many cases, even when they see
substantial procedural evidence. LLM-backed ContractEval has a different error
profile: it tends to preserve omission and ordering sensitivity, while errors
primarily reflect observed-extractor calibration on clean or extra-action cases.

\begin{table*}[t]
\centering
\scriptsize
\resizebox{\textwidth}{!}{%
\begin{tabular}{lrrrrrrr}
\toprule
Gold \ Pred. & Clean & Omit & Branch & Swap & Extra & Inv. & Output \\
\midrule
Clean & 200 & 0 & 0 & 0 & 0 & 0 & 0 \\
Omit & 150 & 50 & 0 & 0 & 0 & 0 & 0 \\
Branch & 0 & 0 & 200 & 0 & 0 & 0 & 0 \\
Swap & 186 & 0 & 0 & 14 & 0 & 0 & 0 \\
Extra & 7 & 0 & 0 & 0 & 193 & 0 & 0 \\
Inv. & 0 & 0 & 0 & 0 & 0 & 197 & 3 \\
Output & 5 & 0 & 0 & 0 & 0 & 0 & 195 \\
\bottomrule
\end{tabular}
}%
\caption{Confusion matrix for Best trace-aware LLM judge (GPT-5.4). Rows are injected labels and columns are predicted labels; each row has 200 cases.}
\label{tab:table_11_confusion_trace_gpt54}
\end{table*}

\begin{table*}[t]
\centering
\scriptsize
\resizebox{\textwidth}{!}{%
\begin{tabular}{lrrrrrrr}
\toprule
Gold \ Pred. & Clean & Omit & Branch & Swap & Extra & Inv. & Output \\
\midrule
Clean & 200 & 0 & 0 & 0 & 0 & 0 & 0 \\
Omit & 157 & 43 & 0 & 0 & 0 & 0 & 0 \\
Branch & 0 & 0 & 200 & 0 & 0 & 0 & 0 \\
Swap & 185 & 0 & 0 & 15 & 0 & 0 & 0 \\
Extra & 4 & 0 & 0 & 0 & 196 & 0 & 0 \\
Inv. & 1 & 0 & 0 & 0 & 0 & 198 & 1 \\
Output & 1 & 0 & 0 & 0 & 0 & 0 & 199 \\
\bottomrule
\end{tabular}
}%
\caption{Confusion matrix for Structured expected-graph LLM judge (GPT-5.4). Rows are injected labels and columns are predicted labels; each row has 200 cases.}
\label{tab:table_12_confusion_expected_graph_gpt54}
\end{table*}

\begin{table*}[t]
\centering
\scriptsize
\resizebox{\textwidth}{!}{%
\begin{tabular}{lrrrrrrr}
\toprule
Gold \ Pred. & Clean & Omit & Branch & Swap & Extra & Inv. & Output \\
\midrule
Clean & 175 & 25 & 0 & 0 & 0 & 0 & 0 \\
Omit & 0 & 200 & 0 & 0 & 0 & 0 & 0 \\
Branch & 0 & 20 & 180 & 0 & 0 & 0 & 0 \\
Swap & 0 & 17 & 0 & 183 & 0 & 0 & 0 \\
Extra & 0 & 29 & 0 & 0 & 171 & 0 & 0 \\
Inv. & 0 & 18 & 0 & 0 & 0 & 182 & 0 \\
Output & 0 & 27 & 0 & 0 & 0 & 0 & 173 \\
\bottomrule
\end{tabular}
}%
\caption{Confusion matrix for ContractEval + GPT-5.4 observed extractor. Rows are injected labels and columns are predicted labels; each row has 200 cases.}
\label{tab:table_13_confusion_contracteval_gpt54_observed}
\end{table*}

\end{document}